# Centroid-aware feature recalibration for cancer grading in pathology images

Jaeung Lee, Keunho Byeon, and Jin Tae Kwak

School of Electrical Engineering, Korea University, Seoul, Republic of Korea
jkwak@korea.ac.kr

**Abstract.** Cancer grading is an essential task in pathology. The recent developments of artificial neural networks in computational pathology have shown that these methods hold great potential for improving the accuracy and quality of cancer diagnosis. However, the issues with the robustness and reliability of such methods have not been fully resolved yet. Herein, we propose a centroid-aware feature recalibration network that can conduct cancer grading in an accurate and robust manner. The proposed network maps an input pathology image into an embedding space and adjusts it by using centroids embedding vectors of different cancer grades via attention mechanism. Equipped with the recalibrated embedding vector, the proposed network classifiers the input pathology image into a pertinent class label, i.e., cancer grade. We evaluate the proposed network using colorectal cancer datasets that were collected under different environments. The experimental results confirm that the proposed network is able to conduct cancer grading in pathology images with high accuracy regardless of the environmental changes in the datasets.

**Keywords:** cancer grading, attention, feature calibration, pathology.

## 1    Introduction

Globally, cancer is a leading cause of death and the burden of cancer incidence and mortality is rapidly growing [1]. In cancer diagnosis, treatment, and management, pathology-driven information plays a pivotal role. Cancer grade is, in particular, one of the major factors that determine the treatment options and life expectancy. However, the current pathology workflow is sub-optimal and low-throughput since it is, by and large, manually conducted, and the large volume of workloads can result in dysfunction or errors in cancer grading, which have an adversarial effect on patient care and safety [2]. Therefore, there is a high demand to automate and expedite the current pathology workflow and to improve the overall accuracy and robustness of cancer grading.

Recently, many computational tools have shown to be effective in analyzing pathology images [3]. These are mainly built based upon deep convolutional neural networks (DCNNs). For instance, [4] used DCCNs for prostate cancer detection and grading, [5] classified gliomas into three different cancer grades, and [6] utilized an ensemble of DCNNs for breast cancer classification. To further improve the efficiency and effectiveness of DCNNs in pathology image analysis, advanced methods that are tailored to pathology images have been proposed. For example, [7] proposed to incorporate both local and global contexts through the aggregation learning of multiple context blocks for colorectal cancer classification; [8] extracted and utilized multi-scale patterns for cancer grading in prostate and colorectal tissues; [9] proposed to re-formulate cancer classification in pathology images as both categorical and ordinal classification problems. Built based upon a shared feature extractor, a categorical classification branch, and an ordinal classification branch, it simultaneously conducts both categorical and ordinal learning for colorectal and prostate cancer grading; a hybrid method that combines DCCNs with hand-crafted features was developed for mitosis detection in breast cancer [10]. Moreover, attention mechanisms have been utilized for an improved pathology image analysis. For instance, [11] proposed a two-step framework for glioma sub-type classification in the brain, which consists of a contrastive learning framework for robust feature extractor training and a sparse-attention block for meaningful multiple instance feature aggregation. Such attention mechanisms have been usually utilized in a multiple instance learning framework or as self-attention for feature representations. To the best of our knowledge, attention mechanisms have not been used for feature representations of class centroids.

In this study, we propose a centroid-aware feature recalibration network (*CaFeNet*) for accurate and robust cancer grading in pathology images. *CaFeNet* is built based upon three major components: 1) a feature extractor, 2) a centroid update (*Cup*) module, and 3) a centroid-aware feature recalibration (*CaFe*) module. The feature extractor is utilized to obtain the feature representation of pathology images. *Cup* module obtains and updates the centroids of class labels, i.e., cancer grades. *CaFe* module adjusts the input embedding vectors with respect to the class centroids (i.e., training data distribution). Assuming that the classes are well separated in the feature space, the centroid embedding vectors can serve as reference points to represent the data distribution of the training data. This indicates that the centroid embedding vectors can be used to recalibrate the input embedding vectors of pathology images. During inference, we fix the centroid embedding vectors so that the recalibrated embedding vectors do not vary much compared to the input embedding vectors even though the data distribution substantially changes, leading to improved stability and robustness of the feature representation. In this manner, the feature representations of the input pathology images are recalibrated and stabilized for a reliable cancer classification. The experimental results demonstrate that *CaFeNet* achieves the state-of-the-art cancer grading performance in colorectal cancer grading datasets. The source code of *CaFeNet* is available at https://github.com/colin19950703/CaFeNet.

## 2 Methodology

The overview of the proposed *CaFeNet* is illustrated in **Fig. 1**. *CaFeNet* employs a deep convolutional neural network as a feature extractor and an attention mechanism to produce robust feature representations of pathology images and conducts cancer grading with high accuracy. **Algorithm 1** depicts the detailed algorithm of *CaFeNet*.

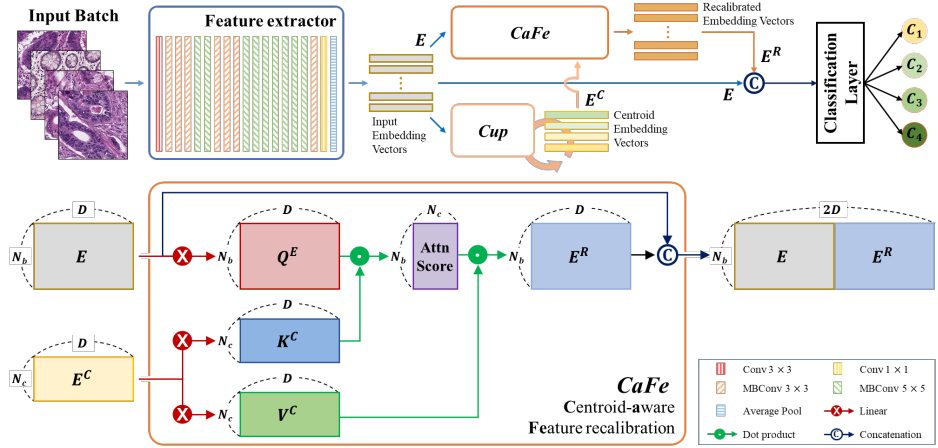

**Fig. 1.** Overview of *CaFeNet*. CaFeNet consists of a feature extractor, a *CaFe* module, a *Cup* module, and a classification layer.

### 2.1 Centroid-aware Feature Recalibration

Let $\{x_i, y_i\}_{i=1}^{N}$ be a set of pairs of pathology images and ground truth labels where $N$ is the number of pathology image-ground truth label pairs, $x_i \in \mathbb{R}^{h \times w \times c}$ is the $i$th pathology image, $y_i \in \{C_1, \ldots, C_M\}$ represents the corresponding ground truth label. $h$, $w$, and $c$ denote the height, width, and the number of channels, respectively. $M$ is the cardinality of the class labels. Given $x_i$, a deep neural network $f$ maps $x_i$ into an embedding space, producing an embedding vector $e_i \in \mathbb{R}^d$. The embedding vector $e_i$ is fed into 1) a centroid update (*Cup*) module and 2) a centroid-aware feature recalibration (*CaFe*) module. *Cup* module obtains and updates the centroid of the class label in the embedding space $E^C \in \mathbb{R}^{M \times D}$. *CaFe* module adjusts the embedding vector $e_i$ in regard to the embedding vectors of the class centroids and produces a recalibrated embedding vector $e_i^{\mathcal{R}}$. $e_i$ and $e_i^{\mathcal{R}}$ are concatenated together and is fed into a classification layer to conduct cancer grading.

Given a batch of input embedding vectors $E = \{e_i | i = 0, \ldots, N-1\}$, *Cup* module computes and updates the centroid embedding vector of each class label per epoch. Specifically, *Cup* module adds up the embedding vectors of different class labels over the iterations per epoch, computes the average embedding vectors, and updates the centroid embedding vectors $E^C = \{e_j^c | j = 0, \ldots, M-1\}$.

*CaFe* module receives a batch of embedding vectors $E = \{e_i | i = 0, \ldots, N-1\}$ and the ground truth labels $Y = \{y_i | i = 0, \ldots, N-1\}$ and a set of centroid embedding vectors $E^C = \{e_j^c | j = 0, \ldots, M-1\}$ and outputs a batch of recalibrated embedding vectors $E^{\mathcal{R}} = \{e_i^{\mathcal{R}} | i = 0, \ldots, N-1\}$ via an attention mechanism (Fig. 1). It first produces queries $Q^E \in \mathbb{R}^{N \times D}$ from $E$ and keys $K^C \in \mathbb{R}^{M \times D}$ and values $V^C \in \mathbb{R}^{M \times D}$ from the centroid embedding vectors $E^C$ by using a linear layer. Then, attention scores are computed via a dot product between $Q^E$ and $K^C$ followed by a softmax operation. Multiplying the attention scores by $V^{\mathcal{C}}$, we obtain the recalibrated feature representation $E^{\mathcal{R}}$ for the input embedding vectors $E$. The process can be formulated as follows:

$$E^{\mathcal{R}} = softmax\left(Q^E K^{\mathcal{C}^T}\right) V^{\mathcal{C}}. \qquad (1)$$

Finally, *CaFe* concatenates $E$ and $E^{\mathcal{R}}$ and produces them as the output.

---
**Algorithm 1** *CaFeNet*
---
**Inputs:** A set of images with ground truth labels $X = \{X^b | b = 1, \ldots, B\}$, $Y = \{Y^b | b = 1, \ldots, B\}$ and current centroid embedding vectors $E^{\mathcal{C}} = \{e_j^{\mathcal{C}} | j = 1, \ldots, N_c\}$
**Outputs:** Predictions $\hat{Y} = \{\hat{Y}^b | b = 1, \ldots, B\}$
1: Initialize next centroids vectors $\tilde{E}^{\mathcal{C}}$ for $j = 1, \ldots, N_c$:
2: $\quad \tilde{e}_j^{\mathcal{C}} \leftarrow 0 \in \mathbb{R}^{1 \times D}$, $S_j \leftarrow 0$
3: **for** $b = 1, \ldots, B$ **do**
4: $\quad$ Take a batch of images and ground truth labels:
5: $\quad\quad X^b \leftarrow \{x_i | i = 1, \ldots, N_b\}$, $x_i \in \mathbb{R}^{h \times w \times c}$
6: $\quad\quad Y^b \leftarrow \{y_i | i = 1, \ldots, N_b\}$, $y_i \in \mathbb{N}$
7: $\quad$ // *CaFe* module computes recalibrated embedding vectors
8: $\quad$ Compute embedding vectors $E = \{e_i | i = 1, \ldots, N_b\}$:
9: $\quad\quad E \leftarrow f(X^b) \in \mathbb{R}^{N_b \times D}$ $\quad\triangleright f$ is a feature extractor
10: $\quad$ Compute query vectors from $E$: $Q^E \leftarrow Linear(E) \in \mathbb{R}^{N_b \times D}$
11: $\quad$ Compute key and value vectors from $E^{\mathcal{C}} = \{e_j^{\mathcal{C}} | j = 1, \ldots, N_c\}$:
12: $\quad\quad K^{\mathcal{C}} \leftarrow Linear(E^{\mathcal{C}}) \in \mathbb{R}^{N_c \times D}$, $V^{\mathcal{C}} \leftarrow Linear(E^{\mathcal{C}}) \in \mathbb{R}^{N_c \times D}$
13: $\quad$ Compute attention scores: $AttnScore \leftarrow Q^E K^{\mathcal{C}^T}$
14: $\quad$ Compute recalibrated vectors $E^{\mathcal{R}} = \{e_j^{\mathcal{R}} | j = 1, \ldots, N_b\}$:
15: $\quad\quad E^{\mathcal{R}} \leftarrow softmax(AttnScore)V^{\mathcal{C}} \in \mathbb{R}^{N_b \times D}$
16: $\quad$ Concatenate two embedding vectors: $E' \leftarrow CONCAT[E, E^{\mathcal{R}}]$
17: $\quad$ Conduct predictions: $\hat{Y}^b \leftarrow Linear(E')$
18: $\quad$ // *Cup* module adds input embedding vectors to next centroids vectors
19: $\quad$ **for** $i = 1, \ldots, N_b$ **do**
20: $\quad\quad$ **for** $j = 1, \ldots, N_c$ **do**
21: $\quad\quad\quad$ **if** $y_i = j$ **then**
22: $\quad\quad\quad\quad$ Add up centroids vectors: $\tilde{e}_j^{\mathcal{C}} \leftarrow \tilde{e}_j^{\mathcal{C}} + e_i$
23: $\quad\quad\quad\quad S_j \leftarrow S_j + 1$
24: $\quad\quad\quad$ **end if**
25: $\quad\quad$ **end for**
26: $\quad$ **end for**
27: $\quad$ // *Cup* module updates centroids embedding vectors
28: $\quad$ **for** $j = 1, \ldots, N_c$ **do**
29: $\quad\quad$ Compute average centroid vectors: $\tilde{e}_j^{\mathcal{C}} \leftarrow \tilde{e}_j^{\mathcal{C}} / S_j$
30: $\quad$ **end for**
31: $\quad$ Update centroid vectors: $E^{\mathcal{C}} \leftarrow \tilde{E}^{\mathcal{C}}$
---

## 2.2 Network Architecture

We employ EfficientNet-B0 [12] as a backbone network. EfficientNet is designed to achieve the state-of-the-art accuracy on computer vision tasks while minimizing computational costs through a compound scaling method. EfficientNet-B0 is composed of one convolution layer and 16 stages of mobile inverted bottleneck blocks, of which each with a different number of layers and channels. Each mobile inverted bottleneck block comprises one pointwise convolution (1×1 convolution for the channel expansion), one depth-wise separable convolution with a kernel size of 3 or 5, and one project pointwise convolution (1×1 convolution for the channel reduction).

Table 1. Details of colorectal cancer datasets

| Class | $C_{Train}$ | $C_{Validation}$ | $C_{TestI}$ | $C_{TestII}$ |
|---|---|---|---|---|
| Benign | 773 | 374 | 453 | 27896 |
| WD | 1866 | 264 | 192 | 8394 |
| MD | 2997 | 370 | 738 | 61985 |
| PD | 1391 | 234 | 205 | 11895 |

## 3 Experiments and Results

### 3.1 Datasets

Two publicly available colorectal cancer datasets [9] were employed to evaluate the effectiveness of the proposed *CaFeNet*. **Table 1** shows the details of the datasets. Both datasets provide colorectal pathology images with ground truth labels for cancer grading. The ground labels are benign (BN), well-differentiated (WD) cancer, moderately-differentiated (MD) cancer, and poorly-differentiated (PD) cancer. The first dataset includes 1600 BN, 2322 WD, 4105 MD, and 1830 PD image patches that were collected between 2006 and 2008 using an Aperio digital slide scanner (Leica Biosystems) at 40x magnification. Each image patch has a spatial size of 1024 × 1024 pixels. This dataset is divided into a training dataset ($C_{Train}$), validation dataset ($C_{Validation}$), and a test dataset ($C_{TestI}$). The second dataset, designated as $C_{TestII}$, contains 27986 BN, 8394 WD, 61985 MD, and 11985 PD image patches of size 1144 × 1144 pixels. These were acquired between 2016 and 2017 using a NanoZoomer digital slide scanner (Hamamatsu Photonics K.K).

### 3.2 Comparative Experiments

We conducted a series of comparative experiments to evaluate the effectiveness of *CaFeNet* for cancer grading, in comparison to several existing methods: 1) three DCNN-based models: ResNet [13], DenseNet [14], EfficientNet [12], 2) two metric learning-based models: triplet loss (Triplet) [15] and supervised contrastive loss (SC) [16], 3) two transformer-based models: vision transformer (ViT) [17] and swin transformer (Swin) [18], and 4) one (pathology) domain-specific model ($\mathcal{M}_{MAE-CE_o}$) [9], which demonstrates the state-of-the-art performance on the two colorectal cancer datasets under consideration. For Triplet and SC, EfficientNet was used as a backbone network. We trained *CaFeNet* and other competing networks on $C_{Train}$ and selected the best model using $C_{Validation}$. Then, the chosen model of each network was separately applied to $C_{TestI}$ and $C_{TestII}$. The results of $\mathcal{M}_{MAE-CE_o}$ were obtained from the original literature.

### 3.3 Implementation Details

We initialized all models using the pre-trained weights on the ImageNet dataset, and then trained them using the Adam optimizer with default parameter values ($\beta_1$ = 0.9, $\beta_2$ = 0.999, $\varepsilon$ = 1.0e-8) for 50 epochs. We employed *cosine anneal warm restart schedule* with initial learning rates of $1.0e^{-3}$, $\eta_{min} = 1.0e^{-3}$, and $T_0 = 20$. After data augmentation, all patches, except for those used in ViT [17] and Swin [18] models, were resized to 512 × 512 pixels. For ViT and Swin, the patches were resized to 384 × 384 pixels. We implemented all models using the PyTorch platform and trained on a workstation equipped with two RTX 3090 GPUs. To increase the variability of the dataset during the training phase, we applied several data augmentation techniques, including affine transformation, random horizontal and vertical flip, image blurring, random Gaussian noise, dropout, random color saturation and contrast conversion, and random contrast transformations. All these techniques were implemented using the Aleju library (https://github.com/aleju/imgaug).

**Table 2.** Result of colorectal cancer grading on $C_{TestI}$.

| Model | Acc (%) | Precision | Recall | F1 | $\kappa_w$ |
|---|---|---|---|---|---|
| ResNet [13] | 87.1 | 0.834 | 0.843 | 0.838 | 0.938 |
| DenseNet [14] | 86.2 | 0.823 | 0.839 | 0.829 | 0.929 |
| EfficientNet [12] | 82.2 | 0.794 | 0.811 | 0.802 | 0.873 |
| Triplet [15] | 86.6 | 0.832 | 0.824 | 0.827 | 0.937 |
| SC [16] | 85.0 | 0.817 | 0.812 | 0.812 | 0.920 |
| ViT [17] | 85.8 | 0.818 | 0.813 | 0.815 | 0.934 |
| Swin [18] | 87.4 | 0.847 | 0.820 | 0.832 | 0.941 |
| $\mathcal{M}_{MAE-CE_o}$ [9] | 87.7 | - | - | 0.843 | 0.940 |
| *CaFeNet* (Ours) | 87.5 | 0.853 | 0.816 | 0.832 | 0.940 |

### 3.4 Result and Discussions

We evaluated the performance of colorectal cancer grading by the proposed *CaFeNet* and other competing models using five evaluation metrics, including accuracy (Acc), precision, recall, F1-score (F1), and quadratic weighted kappa ($\kappa_w$). **Table 2** demonstrates the quantitative experimental results on $C_{TestI}$. The results show that *CaFeNet* was one of the best performing models along with ResNet, Swin, and $\mathcal{M}_{MAE-CE_o}$. were the best performing models. Among DCNN-based models, ResNet was superior to other DCNN-based models. Metric learning was able to improve the classification performance. EffcientNet was the worst model among them, but with the help of triplet loss (Triplet) or supervised contrastive loss (SC), the overall performance increased by ≥2.8% Acc, ≥0.023 precision, ≥0.001 recall, ≥0.010 F1, and ≥0.047 $\kappa_w$. Among the transformer-based models, Swin was one of the best performing models, but ViT showed much lower performance in all evaluation metrics.

**Table 3.** Result of colorectal cancer grading on $C_{TestII}$.

| Model | Acc (%) | Precision | Recall | F1 | $\kappa_w$ |
|---|---|---|---|---|---|
| ResNet [13] | 77.2 | 0.691 | 0.800 | 0.713 | 0.869 |
| DenseNet [14] | 78.8 | 0.698 | 0.792 | 0.722 | 0.866 |
| EfficientNet [12] | 79.3 | 0.701 | 0.802 | 0.727 | 0.870 |
| Triplet [15] | 79.1 | 0.702 | 0.815 | 0.730 | 0.886 |
| SC [16] | 79.7 | 0.718 | 0.809 | 0.739 | 0.876 |
| ViT [17] | 80.7 | 0.706 | 0.797 | 0.733 | 0.889 |
| Swin [18] | 78.6 | 0.690 | 0.785 | 0.712 | 0.873 |
| $\mathcal{M}_{MAE-CE_o}$ [9] | 80.3 | - | - | 0.744 | 0.891 |
| *CaFeNet* (Ours) | 82.7 | 0.728 | 0.810 | 0.756 | 0.901 |

Moreover, we applied the same models to $C_{TestII}$ to test the generalizability of the models. We note that $C_{TestI}$ originated from the same set with $C_{Train}$ and $C_{Validation}$ and $C_{TestII}$ was obtained from different time periods and using a different slide scanner. **Table 3** depicts the quantitative classification results on $C_{TestII}$. *CaFeNet* outperformed other competing models in all evaluation metrics except Triplet for recall. In a head-to-head comparison of the classification results between $C_{TestI}$ and $C_{TestII}$, there was a consistent performance drop in the proposed *CaFeNet* and other competing models. This is ascribable to the difference between the test datasets ($C_{TestI}$ and $C_{TestII}$) and the training and validation datasets ($C_{Train}$ and $C_{Validation}$). In regard to such differences, it is striking that the proposed *CaFeNet* achieved the best performance on $C_{TestII}$. *CaFeNet*, ResNet, Swin, and $\mathcal{M}_{MAE-CE_o}$ were the four best performing models on $C_{TestI}$. However, ResNet, Swin, and $\mathcal{M}_{MAE-CE_o}$ showed a higher performance drop in all evaluation metrics. *CaFeNet* had a minimal performance drop except EfficientNet. EfficientNet, however, obtained poorer performance on both $C_{TestI}$ and $C_{TestII}$. These results suggest that *CaFeNet* has the better generalizability so as to well adapt to unseen histopathology image data.

We conducted ablation experiments to investigate the effect of the *CaFe* module on cancer classification. The results are presented in **Table 4**. The exclusion of the *CaFe* module, i.e., EfficientNet, resulted in much worse performance than *CaFeNet*. Using only the recalibrated embedding vectors $E^{\mathcal{R}}$, a substantial drop in performance was observed. These two results indicate that the recalibrated embedding vectors complement to the input embedding vectors $E$. Moreover, we examined the effect of the method that merges the two embedding vectors. Using addition, instead of concatenation, there was a consistent performance drop, indicating that concatenation is the superior approach for combining the two embedding vectors together.

Table 4. Ablation study on *CaFeNet*.

| Data | Model | Acc (%) | Precision | Recall | F1 | $\kappa_w$ |
|---|---|---|---|---|---|---|
| $C_{TestI}$ | Backbone (EfficientNet) | 82.2 | 0.794 | 0.811 | 0.802 | 0.873 |
| | $E^\mathcal{R}$ only | 77.8 | 0.572 | 0.634 | 0.600 | 0.835 |
| | $ADD[E, E^\mathcal{R}]$ | 82.9 | 0.781 | 0.803 | 0.788 | 0.846 |
| | $CONCAT[E, E^\mathcal{R}]$ (Ours) | 87.5 | 0.853 | 0.816 | 0.832 | 0.940 |
| $C_{TestII}$ | Backbone (EfficientNet) | 79.3 | 0.701 | 0.802 | 0.727 | 0.870 |
| | $E^\mathcal{R}$ only | 56.2 | 0.399 | 0.428 | 0.296 | -0.114 |
| | $ADD[E, E^\mathcal{R}]$ | 75.2 | 0.674 | 0.775 | 0.688 | 0.789 |
| | $CONCAT[E, E^\mathcal{R}]$ (Ours) | 82.7 | 0.728 | 0.810 | 0.756 | 0.901 |

In addition, we compared the model complexity of the proposed *CaFeNet* and other competing models. **Table 5** demonstrates the number of parameters, floating point operations per second (FLOPs), and training and inference time (in milliseconds). The proposed *CaFeNet* was one of the models that require a relatively small number of parameters and FLOPs and a short amount of time during training and inference. Dense-Net, EfficientNet, Triplet, SC, and $\mathcal{M}_{MAE-CE_o}$ contain the smaller number of parameters than that of *CaFeNet*, but these models show either the higher number of FLOPs or longer time during training and/or inference. Similar observations were made for ResNet, ViT, and Swin. These models require much larger number of parameters and FLOPs and longer training time. These results confirm that the proposed *CaFeNet* is computational efficient and it does not achieve its superior learning capability and generalizability at the expense of the model complexity.

Table 5. Model complexity of *CaFeNet* and competing models.

| Model | # Params (M) | # FLOPs (M) | Training (ms/batch) | Inference (ms/batch) |
|---|---|---|---|---|
| ResNet [13] | 23.5 | 21,586.0 | 826.1 | 370.9 |
| DenseNet [14] | 6.7 | 14,802.9 | 1209.9 | 446.2 |
| EfficientNet [12] | 4.0 | 141.1 | 609.8 | 306.4 |
| Triplet [15] | 4.0 | 141.1 | 623.8 | 315.2 |
| SC [16] | 4.0 | 141.1 | 1149.5 | 153.2 |
| ViT [17] | 86.2 | 49,391.9 | 682.4 | 228.8 |
| Swin [18] | 87.3 | 44,659.6 | 1416.14 | 241.8 |
| $\mathcal{M}_{MAE-CE_o}$ [9] | 4.0 | 141.1 | 2344.0 | 519.1 |
| *CaFeNet* (Ours) | 8.9 | 155.9 | 622.6 | 302.9 |

## 4       Conclusions

Herein, we propose an attention mechanism-based deep neural network, called *CaFeNet*, for cancer classification in pathology images. The proposed approach proposes to improve the feature representation of deep neural networks by re-calibrating input embedding vectors via an attention mechanism in regard to the centroids of cancer grades. In the experiments on colorectal cancer datasets against several competing models, the proposed network demonstrated that it has a better learning capability as well as a generalizability in classifying pathology images into different cancer grades. However, the experiments were only conducted on two public colorectal cancer datasets from a single institute. Additional experiments need to be conducted to further verify the findings of our study. Therefore, future work will focus on validating the effectiveness of the proposed network for other types of cancers and tissues in pathology images.

**Acknowledgements**. This work was supported by the National Research Foundation of Korea (NRF) grant funded by the Korea government (MSIT) (No. 2021R1A2C2014557 and No. 2021R1A4A1031864).